\documentclass[review]{elsarticle}

\usepackage{lineno,hyperref}
\usepackage{amsmath}
\usepackage{etoolbox}
\usepackage{lineno}
\usepackage{hyperref}
\modulolinenumbers[5]

\journal{Expert Systems with Applications}

\biboptions{authoryear}

\patchcmd{\pprintMaketitle}
 {\hrule}
 {\clearpage\hrule}
 {}{}
\appto\endfrontmatter{\clearpage}
\begin{document}

\begin{frontmatter}

\title{Machine Learning Approaches for Non-Intrusive Home Absence Detection Based on Appliance Electrical Use. \tnoteref{mytitlenote}}
\tnotetext[mytitlenote]{This research has been co‐financed by the European  Regional Development  Fund  of  the European  Union and Greek national funds through the Operational Program Competitiveness, Entrepreneurship and Innovation, under the call RESEARCH–CREATE–INNOVATE (project code:95699 - Energy Controlling Voice Enabled Intelligent Smart Home Ecosystem)}

\author[1]{Athanasios Lentzas  \cormark[cor1]}
\ead{alentzas@csd.auth.gr}

\author[1]{Dimitris Vrakas}
\ead{dvrakas@csd.auth.gr}

\address[1]{School of Informatics, Aristotle University of Thessaloniki, Thessaloniki, Greece}

\cortext[cor1]{Corresponding author}

\begin{abstract}
Home absence detection is an emerging field on smart home installations. Identifying whether or not the residents of the house are present, is important in numerous scenarios. Possible scenarios include but are not limited to: elderly people living alone, people suffering from dementia, home quarantine. The majority of published papers focus on either pressure / door sensors or cameras in order to detect outing events. Although the aforementioned approaches provide solid results, they are intrusive and require modifications for sensor placement. In our work, appliance electrical use is investigated as a means for detecting the presence or absence of residents. The energy use is the result of power disaggregation, a non intrusive / non invasive sensing method. Since a dataset providing energy data and ground truth for home absence is not available, artificial outing events were introduced on the UK-DALE dataset, a well known dataset for Non Intrusive Load Monitoring (NILM). Several machine learning algorithms were evaluated using the generated dataset. Benchmark results have shown that home absence detection using appliance power consumption is feasible.
\end{abstract}


\end{frontmatter}

\section{Introduction}
Smart homes is an emerging field with increased scientific interest. As ambient sensing devices become cheaper and easily available, smart houses exploit those sensors to acquire information from the surrounding environment \citep{CHAN200855, Nazmiye}. Ambient sensing could be distinguished between obtrusive and unobtrusive sensing. While obtrusive sensing, such as cameras, microphones etc., could potentially provide a better insight, there are privacy concerns especially when data are processed on the cloud.

The smart home concept could promote independent living on elderly people. Activity recognition and abnormal behavior detection can be performed using non intrusive sensing \citep{Lentzas2020}. This is important as signs of mental illnesses (i.e. dementia, Alzheimer's disease) and potentially dangerous conditions, such as falling, can be detected, resulting on immediate intervention.

In addition to abnormal behavior detection, the ability to detect outing events on elderly people living alone is important. Although wandering is a common behavior of people suffering from dementia \citep{SONG2008318,Lai}, a limited amount of published papers address this problem \citep{Lai}. An approach based on door opening detection was proposed by \cite{Aran2016}. Authors assumed that residents close the door when they leave the apartment and open it when they return. The period between two door events as was registered as an outing event. 

Motion and pressure sensors were also used for outing detection by \cite{Petersen2014}. Motion sensors were placed on each room while pressure sensors on doors. A camera was installed at the entrance allowing the authors to identify when the elder was leaving the house. It is worth mentioning that the camera was only used for gathering the ground truth and not for event detection. Logistic regression was employed to detect home absence. Authors linked prolonged outing with less loneliness. 

Pyroelectric sensors were also employed for home absence detection \citep{DBLP:journals/jrm/MoriINSMNOS12}. By monitoring the readings from the sensors, the proposed system was able to detect outing events accurately. Both outing time and duration were logged and authors were able to identify abnormal home absence.  

Similar to \cite{DBLP:journals/jrm/MoriINSMNOS12}, a solution based on infrared sensors was proposed by \cite{Suzuki}. An infrared sensor placed on the entrance was exploited to detect outing events. When the entrance sensor was triggered a going out event was registered. A second sensor trigger after the going out event was indicating that the tenant was back home.

Non intrusive load monitoring (NILM), could provide useful information about the activities performed by the occupants of the house and their behavior \citep{GRAMHANSSEN201894}. NILM requires minuscule intervention at the house, as it is achieved using a single sensor. Using the disaggregated data as input, not only home absence can be detected but also information about the behavior of the residents can be extracted.

While most most works available in the literature perform energy disaggregation on AC power supply, there is an increased interest in DC power supply as well. A power generation source connected to a DC micro grid could potentially eliminate the power loss when converting AC to DC \citep{Quek1,Schirmer}. Due to the increased DC appliances available, new identification techniques needed as AC disaggregation is not applicable to DC micro grid. K-nearest neighbors \citep{Quek3} and One-direction Convolutional LSTM RNN \citep{Quek2} have been proposed. Both produced excellent results on the energy disaggregation problem.

Our approach exploits disaggregated appliance electrical consumption for outing detection, a novel non intrusive, non invasive approach that has not been thoroughly investigated in the past. Since power consumption is a passive unobtrusive sensing method, there are no concerns regarding privacy. Additionally, elder people are more friendly towards less invasive sensors compared to intrusive or wearable sensors. In this paper a benchmark between several machine learning models that can tackle absence detection based on appliance power use is presented. 

\section{Our Approach}
In our work an outing detection approach based on electricity use of main appliances is proposed. Monitoring the power consumption can be achieved with unobtrusive techniques that require minimum intervention in the house. Smart plugs could be used on wall sockets in order to monitor the electrical use or it could be incorporated in a power disaggregation system, where the disaggregated energy consumption is used as input on the outing detection module.

In this section our motivation, data collection, our methodology and the experiments performed are discussed.

\subsection{Motivation}
Outing detection based on appliance use could be beneficial in many real world scenarios where it is important to know whether or not the residents of a house are present. Our motivation was based on two scenarios: 1) Elderly people living alone and the early detection of mental diseases. 2) Monitor home quarantine especially during pandemics. With the outbreak of COVID-19, specialists recommend home isolation especially for people suspected to be sick. Despite instructions, many of them break their home quarantine helping the spread of corona-virus. An effective, cheap and easy to install home absence detection system would be extremely useful in cases like that. 

Contributions of an approach based on electrical consumption would be significant: a) the system would be easy to deploy without excessive house modifications. b) Load monitoring is a non intrusive method. Personal privacy can also be retained as, given a low enough sample frequency, machine learning algorithms can not extract specific knowledge. c) Houses with similar resident and electrical consumption profile could benefit from pre-trained machine learning models. d) An evaluation of different machine learning techniques would provide a baseline for future research. 

Preserving privacy in a strong motivation on smart home installations that monitor energy use \citep{6194398,Ukil}. As \cite{6194398} mention in their work, privacy issues are regulated by government agencies in most countries with strict laws. Elderly people have a different level of acceptance of information technology monitoring their health compared to young population \citep{FISCHER2014624}. Privacy and social stigmatization are important barriers that must be addressed for health monitoring tools. In addition to that, ethical issues arise when private information are gathered, transmitted and processed \citep{10.1007/978-3-319-57348-9_22}. A possible disclosure of sensitive information, may have a negative impact on the social life of the person, as it could be stigmatized. 

\subsection{Dataset}
The dataset used in our research is the UK Domestic Appliance-Level Electricity (UK-DALE) \citep{UK-DALE}. The dataset contains electrical use from five houses. Both the total consumption as well as disaggregated consumption of individual appliances are recorded. The use was recorded every 6 seconds. Data from house 1 were used in our work, covering a period of 4.3 years.

In our work, four appliances were employed to detect home absence: television, kettle, oven and microwave. As the dataset does not contain annotated data for home absence, the data were manually annotated. Everyday outings (i.e. home absence events) were added based on the consumption of the aforementioned devices. Additionally, major outing events were added. A Christmas trip starting on December $24^{th}$ till $26^{th}$, a 4-days spring break randomly placed in the period between $15^{th}$ of March and end of May, summer vacation for 2 weeks on early August and a weekend trip during autumn. The Christmas trip could be extended till $28{th}$ in case the $26^{th}$ is a weekend. 

In addition to the aforementioned outings, daily outing events were also introduced. Each week day, home absence was added between 8:30 in the morning and 16:00 in the afternoon, in case no electrical use was registered in that time frame. As a result a daily work routine was simulated. Finally, random outing events were added on Saturdays.

An additional binary feature was introduced indicating whether the day was a bank holiday / weekend or not. While our intuition was that such a feature would improve the performance, that did not happen. The aforementioned binary feature was excluded in order to avoid overfitting (e.g. assume that people are always away on bank holidays). 

\begin{figure}
    \centering
    \includegraphics[width=1\textwidth]{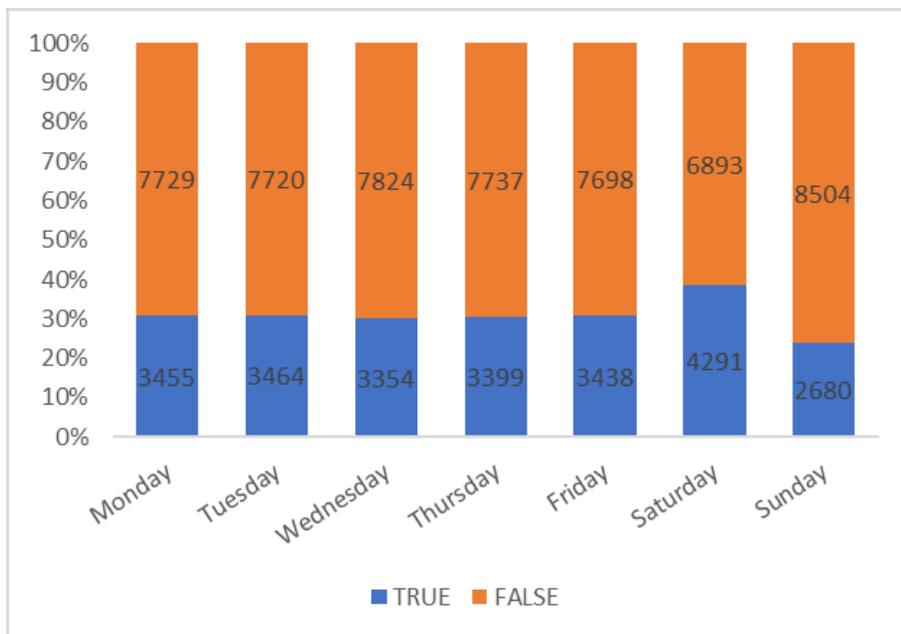}
    \caption{Daily Outing events distribution}
    \label{fig:outing}
\end{figure}

As already mentioned, the dataset was over-sampled with measurements taken every 6 seconds. Since detecting home absence in such a small time window is futile, the dataset was down-sampled. Measures were re-sampled to 30 minutes and the mean of all disaggregated values was used. The result was a dataset with 78.186 total instances, where 24.081 were outing events. The distribution of outing events on each weekday can be seen on Figure \ref{fig:outing}.

One of the main reasons we had to create our own data was the absence of a publicly available electric use dataset with home absence reported (based on our research). Although there are some datasets with outing ground truth, such as the Intelligent Systems for Assessing Aging Changes (ISAAC) \citep{Kaye} cohort and the ORCATECH Living Lab, the majority of them rely on ambient sensors and cameras. Pressure sensors are usually placed on door mats and / or contact sensors are placed on doors in order to detect when someone leaves or enters a house.

\subsection{Experiments}
Several machine learning algorithms were evaluated. More specifically Decision Tables (DT), C4.5 Trees \citep{Salzberg1994}, Random Forests (RF) \citep{TinRF}, Naive Bayes (NB), Multilayer Perceptron (MLP) \citep{Rosenblatt} and Deep Neural Networks \citep{LeCun2015} (DNN) were employed. Multilayer Perceptron was limited less than 10 hidden layers while Deep Neural Network's hidden layers were left unlimited. All methods were implemented using Python, SciKit Learn \citep{scikit-learn} and Keras \citep{chollet2015keras} with Tensorflow \citep{tensorflow2015-whitepaper} backend. The selection of those algorithms was based on related works found on the literature \citep{Lentzas2020}. Additionally, they are some of the most commonly used machine learning algorithms \citep{7724478}. 

A decision table is a machine learning algorithm documenting the different decisions taken depending on different set of conditions (similar to a flowchart). They still remain a solid choice on machine learning since they have a series of advantages. One of the main advantages of decision tables is that the result is interpretable allowing one to understand the reasoning behind the decision taken. Additionally they are immune to data scaling and multicolinearity as well as less feature engineering is required. The main drawback is the tendency to overfit and the requirement to load all the data to memory (batch training is not supported).

C4.5 tree is a variant of decision trees. It is one of the most widely used machine learning algorithm \citep[p.~191]{WittenFrankHall11}. Information entropy is used to build the tree decision tree using a set of training data. At each node a feature is chosen and the training dataset is split into subsets enriched in one of the classes. Information gain is the used as splitting criterion. Recursive execution over the subsets creates the final tree. In order to avoid overfitting, pruning is used. In our work post-pruning was used. The entire tree was built first and then certain branches were removed using sub-tree replacement (a sub-tree is replaced with a leaf if the classification error is reduced). While C4.5 trees are interpretable and can deal with noise efficiently they require a bigger training set and small variations in data can lead to different trees. 

Random forests, as the name implies, consists of multiple decision trees. Each tree predicts one class and the class that was predicted most times is the output of the model. This ensemble method achieve higher accuracy compared to decisions trees. Interpretability is sacrificed but a higher accuracy is usually achieved (compared to a single decision tree). Accuracy may not increase in case of problems with multiple categorical variables and when there is a linear correlation between the predictive attributes and the target variable \citep{SMITH201385}.  In our work C4.5 trees were used to create the Random Forest. 

Naive Bayes classifiers are based on Bayes theorem and are among the simplest Bayesian network models. They can achieve high accuracy when coupled with kernel density estimation. Kernel density estimation was used in our experiments. NB are faster when compared with more sophisticated methods while achieving good accuracy. 

Multilayer perceptron is a feedforward artificial neural network.  MLP consists of at least three layers of nodes (input, output and one or more hidden). In our work hidden layers were limited to 10. On the other hand, the number of layers on the Deep Neural Network were left unlimited. 

DNN are artificial neural networks with multiple layers between the input and output layers. They can model complex non-linear relationships. Overfitting and computation time is common issues both on DNN and MLP. 

\subsubsection{Hyperparameter Tuning}
All the aforementioned machine learning algorithms have several parameters that require tuning before training. There are several techniques that can help researches tune those parameters (although some of them could be set based on intuition). 

In our work all the models (except from the DNN) had there parameters tuned using quantum genetic algorithms as described by \cite{LentzasQ}. Quantum genetics algorithms (QGA) \citep{Narayanan2002} is a variation of genetic algorithms \citep{Goldberg1979}, an evolutionary type of algorithm. The main advantage of QGA is the faster convergence to the local best while performing global search. Each combination of hyperparameters was expressed as a quantum chromosome. A model was trained for each chromosome and based on their respective performance, the population is evolving, converging to the best solution. In order to avoid local optimal solutions, quantum disaster was employed \citep{Miao2009}. 

DNN hyperparameter tuning was performed with  Randomized Parameter Optimization \citep{Bergstra}. A set or range (in case of continuous variables) is given for every hyperparameter. A random search is then performed on these distributions for a total of $N$ times. For each combination of hyperparameters a model is trained and the hyperparameters associated with the highest accuracy model are returned.  

\subsubsection{Evaluation Metrics}
Evaluating the performance of each algorithm used was based on the most common metrics used in binary classification: Accuracy, Precision, Recall and F-score. In order to calculate those metrics, the confusion matrix is used. Confusion matrix is a $C_{n \times n}$ matrix where $n$ is the number of classes. Each element $C_{i,j}$ is the number of examples that belong to $i_{th}$ class and classified as instances of $j_{th}$ class. Confusion matrix allows the extraction of useful numerical values used on the aforementioned metrics: 
\begin{itemize}
    \item True Positives (TP): positive instances classified correctly.
    \item True Negatives (TN): negative instances classified correctly.
    \item False Positives (FP): negative instances classified as positive
    \item False Negatives (FN): positive instances classified as negatives
\end{itemize}

The information mentioned above are used in metrics calculation. More specifically: \begin{itemize}
    \item Accuracy: percentage of correctly classified examples (both positive and negative)
    \item Precision / Recall: ratio of correctly classified instances to the total positive classified examples and total positive instances respectively. 
    \item F-score (F1): combination of Precision and Recall in one single metric. 
\end{itemize}

\begin{align*}
    Accuracy = \frac{\sum_{TP} + \sum_{TN}}{\sum_{Examples}} && Precision = \frac{\sum_{TP}}{\sum_{TP} + \sum_{FP}}\\
    Recall = \frac{\sum_{TP}}{\sum_{TP} + \sum_{FN}} && F1 = 2\frac{Precision \times Recall}{Precision + Recall}
\end{align*}

\subsection{Evaluation}
Each model was trained using the same features. The chosen features can be seen on Table \ref{features}. Appliance use was converted to a binary variable based on whether the device on or off. Appliances with use lower than 10 Watts were considered off and the rest were set to on. By observing the dataset, when an appliance was operating, the electrical power consumed was greater than 30 Watts, while a device was on standby, the consumption was approximately 1. Setting the threshold to 10 Watts, allowed us to rule out power spikes identified in the data. The weak day variable was encoded to categorical variable in range [0-6] with (0) representing Monday and (6) Sunday. Label (ordinal) encoding \citep{VONEYE1996xix} was preferred over one-hot-encoding \citep{lantz_2015}. In the first place we would like to imply an order on week days. Moreover, using ordinal encoding, we avoided increasing the dimension of our dataset. Lastly, there was not a significant performance loss (as seen on Table \ref{results_average},\ref{results_table} while using label encoding.

\begin{table}[htbp]
    \centering
    \begin{tabular}{c|c}
        \textbf{Feature} & \textbf{Description}  \\
        Appliance on/off & Binary variable indicating whether the appliance was on or off \\
        Time & The time on 24h format \\
        Weak day & The day of the week (Monday - Sunday) \\
        Day & Day part of the date (1-31) \\
        Month & The month of the observation(1-12)
    \end{tabular}
    \caption{Features used}
    \label{features}
\end{table}

During evaluation 10-fold cross validation was employed. Each experiment was repeated 10 times. The results can be seen on Table \ref{results_table}, \ref{results_average}. The former table contains the results of the best run while the latter the average results of all runs.
As one can observe on Table \ref{results_table} MLP had the best single run compared to the rest of the classifiers evaluated. C4.5 trees, Decision Table and Random Forests had performance on par with the MLP. Naive Bayes and Deep NN on the other hand had the worst performance compared to the rest, but still achieved an overall good score.

Observing the average of the 10 runs executed (Table \ref{results_average}) more information about the performance of each model can be extracted. Compared to the best run, averaging the results suggests that C4.5 Trees performance is better. Comparing the average score with the best score achieved, C4.5 and Decision Table had the smaller difference, thus providing more robust results.

\begin{table}[htbp]
    \centering
    \begin{tabular}{c|c|c|c|c}
         \textbf{Classifier} & \textbf{Accuracy} & \textbf{Precision} & \textbf{Recall} & \textbf{F-score}  \\
        Decision Table & 0.9774 & 0.9861 & 0.986 & 0.9847 \\
        C4.5 Tree & 0.9795 & 0.9916 & 0.9842 & 0.9861 \\
        Random Forests & 0.9758 & 0.9829 & 0.9856 & 0.9836  \\
        Naive Bayes & 0.8766 & 0.8894 & 0.9602 & 0.9192 \\
        MLP & \textbf{0.982} & \textbf{0.9927} & \textbf{0.9946} & \textbf{0.9877} \\
        Deep NN & 0.8671 & 0.9035 & 0.9404 & 0.9116
    \end{tabular} 
    \caption{Best Accuracy, Precision, Recall, F-score per classifier}
    \label{results_table}
\end{table}

\begin{table}[htbp]
    \centering
    \begin{tabular}{c|c|c|c|c}
         \textbf{Classifier} & \textbf{Accuracy} & \textbf{Precision} & \textbf{Recall} & \textbf{F-score}  \\
        Decision Table & 0.974 & 0.9819 & \textbf{0.9829} & 0.9824 \\
        C4.5 Tree & \textbf{0.9766} & \textbf{0.9882} & 0.98 & \textbf{0.9841} \\
        Random Forests & 0.9710 & 0.9788 & 0.982 & 0.9804  \\
        Naive Bayes & 0.8678 & 0.8781 & 0.9533 & 0.9141 \\
        MLP & 0.9727 & 0.9824 & 0.9809 & 0.98156 \\
        Deep NN & 0.8575 & 0.8877 & 0.9238 & 0.9054
    \end{tabular} 
    \caption{Average Accuracy, Precision, Recall, F-score per classifier}
    \label{results_average}
\end{table}

Since the performance of most classifiers was similar a paired t-test analysis was performed using Weka \citep{hall09:_weka_data_minin_softw}. The t-test analysis was performed with 0.05 confidence and F-score was used. The best classifier (C4.5) was used as the baseline for the analysis. T-test analysis shown that results obtained from the rest of the models were statistically worst compared to the baseline.

\section{Conclusions \& Future Work}
Outing detection based on appliance use is a field that despite the benefits it could provide is not thoroughly discussed in the literature. This paper provides a benchmark of several machine learning models. UK-Dale dataset was used for training. As no home absence events were present in the dataset, artificial events were introduced.

Results shown that home absence detection based on electrical consumption is feasible. Although MLP had the best score on a single run, C4.5 tree achieved the best average score. T-test analysis shown that C4.5 tree had statistically better results compared to the rest of the benchmarked classifiers. 

Home absence detection could be applied in a variety of scenarios. A potential application is planning home delivery routes. Knowing, or predicting, outing periods, delivery companies could plan their daily routes to deliver as many parcels as possible. This is important since delivering as many goods as possible without having to reschedule for another day, could reduce operating costs.

Smart home applications targeting elder people leaving alone, could also utilize a home absence detection module. Detecting prolonged periods of home absence, especially during abnormal time periods, could lead to early detection of emergencies or mental issues for elders. 

Additionally, elders suffering from dementia could benefit from a smart home able to detect outing events. According to Alzheimer's Association, wandering off could occur in people suffering from dementia. Having the ability to detect whether the patient left the house could lead to an early alarm to his relatives or caretakers. This could promote independent living of dementia patients.

Knowing when a person is out of home could be useful on energy saving systems as well. A smart home installation could take certain actions in order to preserve power and reduce cost and environmental impact when the owners are out of home. For instance a smart thermostat could turn off the heating when the house is empty and turn it on in time before the residents return. A boiler could be automatically turned off when left on during an outing event. All these applications would reduce the energy requirements of a smart house.

Future work could be focused on integrating home absence detection on a power disaggregation module. Machine learning approaches provide a solid power disaggregation method \citep{Nalmpantis2019}. Using the disaggregated results provided by a system deployed on a real house, will provide a complete outing detection system. Further investigation is needed though as disaggregation error could accumulate and impact absence detection.

Transfer learning should also be evaluated. As already mentioned on motivation section, pre-trained models on large datasets could be exploited. Especially on houses with alike energy demand profiles and residents following similar outing patterns. Transfer learning would greatly reduce deployment time without severely affecting performance.

In addition to the above, a non artificially created dataset will be collected. Ground truth could be logged with sensors on the outer door and door mat. This could provide a more robust approach on outing detection based on electrical use. Although the rules applied to the dataset were selected based on accurate simulation of human behavior, the resulting dataset could be biased compared to data collected from a real person. Even by applying a random factor when the dataset was generated, human behavior can't be described precisely by rules. 

\bibliography{mybibfile}

\end{document}